\documentclass[conference]{IEEEtran}
\IEEEoverridecommandlockouts
\usepackage{cite}
\usepackage{amsmath,amssymb,amsfonts}
\usepackage{algorithmic}
\usepackage{xspace}
\usepackage{graphicx}
\usepackage{textcomp}
\usepackage[table,dvipsnames]{xcolor}
\usepackage{booktabs}
\usepackage{hyperref}
\usepackage{enumitem}

\newcommand{\castXIX}{CAsT 2019\xspace}
\newcommand{\castXX}{CAsT 2020\xspace}

\newcommand{\promptnumber}{five\xspace}
\newcommand{\gpt}{\texttt{gpt-3.5-turbo}\xspace}
\usepackage{soul}
\usepackage{xcolor}
\usepackage[dvipsnames]{xcolor}

\newcommand{\guido}[1]{\textcolor{black}{#1}}

\newcommand{\mycomment}[1]{}
\def\BibTeX{{\rm B\kern-.05em{\sc i\kern-.025em b}\kern-.08em
    T\kern-.1667em\lower.7ex\hbox{E}\kern-.125emX}}

\usepackage{microtype}

\usepackage[show]{chato-notes}

\begin{document}

\title{Rewriting Conversational Utterances with\\Instructed Large Language Models}

\author{
\IEEEauthorblockN{Elnara Galimzhanova}
\IEEEauthorblockA{University of Pisa, Pisa, Italy\\e.galimzhanova@studenti.unipi.it}

\and

\IEEEauthorblockN{Cristina Ioana Muntean, Franco Maria Nardini,\\ Raffaele Perego, Guido Rocchietti}
\IEEEauthorblockA{ISTI-CNR, Pisa, Italy\\\{name.surname\}@isti.cnr.it}
\vspace{-25mm}
}

\maketitle

% !TEX root = ../paper.tex
% !TeX spellcheck = en_US

\begin{abstract}
Many recent studies have shown the ability of large language models (LLMs) to achieve state-of-the-art performance on many NLP tasks, such as question answering, text summarization, coding, and translation. In some cases, the results provided by LLMs are on par with those of human experts.
These models' most disruptive innovation is their ability to perform tasks via zero-shot or few-shot prompting.
This capability has been successfully exploited to train \emph{instructed} LLMs, where reinforcement learning with human feedback is used to guide the model to follow the user's requests directly.
In this paper, we investigate the ability of instructed LLMs  to %improve retrieval effectiveness. Specifically, we study the power of an instructed LLM to 
improve conversational search effectiveness by rewriting  user questions in a conversational setting. We study which prompts %, i.e., the interaction template used to interact with the instructed LLM, 
provide the most informative rewritten utterances that lead to the best retrieval performance. %. We conduct our investigation in a specific information retrieval scenario, i.e., conversational search, 
Reproducible experiments are conducted on publicly-available TREC CAST datasets. The results show that rewriting conversational utterances with instructed LLMs achieves significant improvements of up to 25.2\% in MRR,  31.7\% in Precision@1, 27\% in NDCG@3, and 11.5\% in Recall@500 over state-of-the-art techniques.% in conversational search.
\end{abstract}

\begin{IEEEkeywords}
conversational systems, query rewriting, LLMs, ChatGPT, information retrieval
\vspace{-5mm}
\end{IEEEkeywords}

%% The code below is generated by the tool at http://dl.acm.org/ccs.cfm.
%% Please copy and paste the code instead of the example below.
%%
\iffalse


%% Keywords. The author(s) should pick words that accurately describe
%% the work being presented. Separate the keywords with commas.
\fi

%% This command processes the author and affiliation and title
%% information and builds the first part of the formatted document.

\maketitle

% !TEX root = ../paper.tex
% !TeX spellcheck = en_US

\section{Introduction}
\label{sec:introduction}

Since their introduction,  Large Language Models (LLMs) have impressed with their capabilities in dealing with  tasks such as question answering, text  summarization, coding, and translation, with performances that are comparable to those of human annotators.
Thanks to their ability to perform tasks via few-shot learning, LLMs can learn from just a few examples, considerably expanding the range of applications supported and lowering the effort needed for targeting  novel tasks.
This feature has been successfully exploited to train \textit{Instructed} LLMs, where methods from reinforcement learning with human feedback (RLHF) are used to directly instruct the model to act following the user’s intention~\cite{ouyang2022training}.

As a result, we assisted a new gold rush for part of the major tech companies to show their new intelligent systems.
At first, we witness the introduction of ChatGPT, powered by a GPT-3.5 model. Then, we witness the release of a novel version of Bing search powered by GPT-4. 
% \fnote{should we mention also models like: bart, alpaca etc., not just OpenAI?}
The availability of the GPT-4-powered Bing search engine sets a definitive shift from a search paradigm based on ``ten blue links'' returned  as an answer to a user query to a natural-language answer that is then returned to the user. Such an autonomous system automatically chooses the most relevant documents and extracts and elaborates the relevant information that is then presented to the user in the form of an answer to her/his query.  %\fnote{are we sure it works that way? I think it gives a "generated" answer and then runs the query through the search engine to find out also web pages} 
This novel paradigm that exploits the dialogue to interact between the user and the search system can indeed provide a more friendly and natural way of interacting with the search service.

In this paper, we move a step forward in an orthogonal direction by studying the ability of instructed LLMs to improve the retrieval effectiveness of a state-of-the-art search engine in a conversational setting\cite{dalton_cast_2019,dalton_cast_2020,dalton_trec_2021}. 
%\mycomment{Specifically, we address  conversational search  \cite{dalton_cast_2019,dalton_cast_2020,dalton_trec_2021}, aimed to enable users to interact with a search system using natural language, and to receive relevant and informative answers in a conversational format.}
%\mycomment{The task involves understanding the user's information need, retrieving relevant information from various sources, generating natural and coherent responses, and maintaining a dialogue context over multiple turns.  Rhetorical figures -- e.g., anaphoras, ellipses, coreferences -- and complex speech constructs in users' utterances make conversational search challenging. }
We aim to answer two main research questions:

\begin{itemize}[leftmargin=10mm]
    \item[RQ1] Can an instructed LLM improve conversational search effectiveness by automatically rewriting  the users' utterances  to allow the search engine to retrieve more precise and relevant results?
    \item[RQ2] Which prompting template performs best in order to generate rewritten queries that enhance retrieval performance?
\end{itemize}

We investigate the research questions above by adopting the Conversational Assistance Track (CAsT) framework provided by TREC for training and evaluating models in open-domain information-centric conversational dialogues~\cite{dalton_cast_2019}.
% \mycomment{
% The CAsT framework provides us with multi-turn conversations and utterance-level relevance assessments for documents belonging to a publicly-available collection.
% Investigating the two research questions above in conversational search is interesting because, when viewed singularly, utterances in a conversation may lack the \emph{context} emerging from the entire conversation. Moreover, conversations start with a central topic. They may evolve with different facets of the initial topic or an abrupt shift to a new focus, possibly suggested by the content of the answers returned~\cite{mele_topic_2020, mele_adaptive_2021}.}

The characteristics of conversational utterances, i.e., missing context from previous questions, topic shifts~\cite{mele_topic_2020, mele_adaptive_2021}, and implied concepts from previous answers, pose new challenges to deal with, which are a direct consequence of the paradigm shift introduced by conversational search. They heavily impact the performance of standard information retrieval techniques. %when   this scenario. 
Query rewriting techniques applied on a per-utterance level answer these challenges as they help propagate the context throughout the conversation and deal with possible topic shifts.

\iffalse
This gives us the possibility to investigate a further interesting research question:
\begin{itemize}
    \item[RQ3] To what extent an instructed LLM can follow the context of a conversational dialogue, manage the possible topic shifts and propagate the correct contextual information from utterance to utterance? 
\end{itemize}
\fi

The novel contributions of this work are thus the following:
\begin{itemize}
    \item We investigate utterance rewriting in conversational search using an instructed LLM and specifically designed prompting templates. Given an utterance and its context, we prompt %an instructed LLM
    the model asking to generate a rewriting of the utterance with the goal of enhancing the retrieval effectiveness of a state-of-the-art information retrieval system. This approach allows us to evaluate the ability of an instructed LLM to deal with the context of a conversation and possible topic shifts that may occur. At the same time, we inspect its ability to  rewrite  natural language utterances  containing ambiguities, coreferences, omissions, acronyms, and colloquial grammar misuses.
    \item We present \promptnumber different prompting templates to rewrite the utterances. Each prompt has been evaluated in an end-to-end retrieval framework to assess its ability to improve the effectiveness of the conversational search system. All of the prompts have been tested in different conditions to establish the best way of prompting an LLM.
    \item We report the results of a comprehensive and reproducible experimental evaluation conducted using the publicly-available TREC CAsT datasets. Results show that rewriting utterances with the chosen instructed LLM achieves significant improvements of up to 31.7\% in Precision@1, 25.2\% in MRR, 27\% in NDCG@3, and 11.5 \% in Recall@500 over state-of-the-art rewriting techniques in conversational search.
    %The results show that we obtain improvements up to 31.7\% in P@1 and 27\% in NDCG@3 over state-of-the-art baselines.
\end{itemize}

The rest of the paper is organized as follows. 
Section~\ref{sec:related} discusses related work, while Section~\ref{sec:methodology} introduces our methodology. In Section~\ref{sec:expsetup} we discuss the details of the prompting templates designed for query rewriting, the datasets, the baselines and competitors, and the two-stage retrieval  architecture. The end-to-end performance of the proposed query rewriting pipeline is comprehensively assessed in Section~\ref{sec:results}. Finally, Section~\ref{sec:conclusion} presents the concluding remarks.

% !TEX root = ../paper.tex
% !TeX spellcheck = en_US

\section{Related Work}
\label{sec:related}

\paragraph{Conversational search} Query rewriting is central in modern web search as it better models the user's information need and enhances retrieval effectiveness~\cite{he2016learning}.
Similar challenges arise in conversational search, since utterances, like queries, may be ambiguous or poorly formulated.

Conversational utterance rewriting aims to reformulate a concise request in a conversational context to a fully specified, context-independent query dealing with anaphoras, ellipses, and other linguistic phenomena\cite{yang2017neural,mele_topic_2020}.% that information retrieval systems can handle  effectively. 
\mycomment{To achieve that, utterance rewriting techniques focus on handling linguistic characteristics of human dialogues, such as anaphora (words that explicitly reference previous conversation turns) and ellipsis (one or more words that are omitted from the conversation and nevertheless understood in the context of the remaining elements of the conversation) \cite{yang2017neural,mele_topic_2020}.} These techniques aim at identifying terms previously mentioned in the conversation to expand the current utterance profitably \cite{aliannejadi_harnessing_2020,10.1145/3397271.3401130,mele_adaptive_2021, rocchietti2023}. In this line, 
Aliannejadi \emph{et al.} propose a novel neural utterance relevance model based on BERT that helps identify the utterances relevant to a given turn \cite{aliannejadi_harnessing_2020}.
Voskarides \emph{et al.}~\cite{10.1145/3397271.3401130} model query rewriting for conversational search as a binary term classification task and introduce QuReTeC, a Bi-LSTM model that selects the valuable terms in context to enrich the query. \mycomment{For each term in the previous conversation turns, QuReTeC decides whether to add it to the current query or not. The model encodes the conversation history and the current query using BERT and uses a term classification layer to predict a binary label for each term in the conversation history.}

Other approaches rewrite the utterances by exploiting a fine-tuned neural model \cite{10.1145/3397271.3401323,Hao2022,vakulenko_question_2021,su_improving_2019}. 
Yu \emph{et al.} presents CQR, a few-shot generative approach to solve coreference and omissions in conversational query rewriting \cite{10.1145/3397271.3401323}. The authors propose two methods to solve coreference and omissions to generate weak supervision data that are then used to fine-tune GPT-2 to rewrite conversational queries. Results show that on the TREC CAsT Track a weakly-supervised finetuning of GPT-2 improves the ranking accuracy by 12\%. 

Vakulenko \emph{et al.}~\cite{vakulenko_question_2021} approach the problem by tackling conversational question answering. The authors propose a question-rewriting technique that translates ambiguous requests into semantically-equivalent unambiguous questions. \mycomment{The question-rewriting model employs a unidirectional Transformer decoder whose input is the (potentially ambiguous) current question plus five previous conversation turns. The model is trained using a set of ground-truth questions manually rewritten by human annotators. Given a sequence of previous tokens, the model predicts the next token in an output sequence.}

%commentato
\mycomment{
In a follow-up work, Vakulenko \emph{et al.} \cite{vakulenko_comparison_2021} compare original user questions and human-rewritten questions against questions automatically rewritten by sequence-generation models based on GPT-2 and by QuReTeC. The authors also prove that combining different models can improve performance. In particular, simply appending the terms predicted by QuReTeC to the questions rewritten by a sequence-generation model improve the state-of-the-art ranking performance.}

In more recent works, several papers exploit pre-trained language models to represent queries and documents in the same dense latent vector space and then use the inner product to compute the relevance score of a document to a given query. 
In conversational search, the representation of a query can be computed in two different ways. In one case, a stand-alone contextual query understanding module reformulates the user query into a rewritten query, exploiting the context history~\cite{DBLP:journals/corr/abs-2201-05176}, and then a query embedding is computed, e.g. using embedding models such as ANCE\cite{ance-paper} or STAR\cite{star-paper}. Alternatively, the learned representation function is trained to receive as input the query together with its context history and to generate a query embedding that is more similar to the manual query embeddings~\cite{10.1145/3404835.3462856}. In both cases, dense retrieval methods are used to compute the query-document similarity by deploying efficient nearest neighbor techniques over specialized indexes, such as those provided by the FAISS toolkit~\cite{JDH17}.

\paragraph{Large Language Models}
 LLMs based on transformer architectures such as GPT are trained on large corpora of text data to comprehend and produce natural language~\cite{Radford,Vaswani}. The pre-trained models produced with unsupervised  training~\cite{brown2020language} can be easily fine-tuned for various tasks in a supervised setting.  InstructGPT, based on GPT-3, has been fine-tuned using human feedback to make it better at following user intentions \cite{ouyang2022training}. %In contrast to GPT-3, InstructGPT models are less likely to produce false or harmful outputs. 
 Bidirectional and Auto-Regressive Transformer (BART) integrate the strengths of two established models, i.e., BERT and GPT-2, and are trained using a denoising autoencoder approach to understand text structure and semantics, as well as generate fluent and coherent text~\cite{lewis2019bart}. 
%Taori \emph{et al.} ~\cite{Taori2023alpaca} presented discoveries on a language model called Alpaca that is designed to follow instructions. Alpaca is based on Meta's LLaMA 7B model and was fine-tuned using 52,000 demonstrations of instruction-following that were generated in the style of self-instruct with the \texttt{text-davinci-003} algorithm ~\cite{Taori2023alpaca}. 
Another instructed LLM model of the GPT family is ChatGPT\footnote{\url{https://chat.openai.com/}}, which  is explicitly tailored for conversational applications~\cite{liu2023summary}.  

Instructed LLMs such as ChatGPT are easily adaptable to new tasks and domains, making them very useful in various tasks. Wei \emph{et al.}~ \cite{wei2023zeroshot} propose ChatIE, a framework that employs ChatGPT to perform zero-shot Information Extraction (IE) tasks via multi-turn question-answering and claim that their method can achieve impressive results and surpass some full-shot models across three IE tasks. 
%commentato
% \mycomment{
% Additionally, the effectiveness of ChatGPT in generating high-precision and high-recall Boolean queries for systematic review literature search is explored by Wang \emph{et al.}~\cite{wang2023chatgpt}.
% Dai {et al.}~\cite{dai2023uncovering} investigates the performance of ChatGPT in generating recommendations for different domains and scenarios, showing that it can outperform other large language models and handle the cold start problem while providing interpretable recommendations.}
Sun {et al.}~\cite{sun2023chatgpt} found that ChatGPT can perform as well as, or better than, supervised methods in information retrieval relevance ranking when guided by domain-specific guidelines. The models mentioned earlier achieve impressive results in many NLP tasks, and their applications are many, from medicine to finance and beyond. With proper instructions, these models can solve a vast variety of tasks, making them valuable tools for researchers and developers alike.
ChatGPT is  the instructed LLM we use in  our experiments.

%A  paper that goes in a similar direction as ours related to ours 
\guido{Lately
%is the one of
Mao et al.~\cite{mao2023large} conducted a work that studies the impact of LLMs}. They focus on capturing the contextual conversational search intent through the use of  %\texttt{code-davinci-002}
GPT-3. The authors evaluate their findings in an ad-hoc dense retrieval scenario, using ANCE embeddings \cite{ance-paper} for computing the similarity scores between documents and queries. We use their best-performing prompt in our experimental setting to see its effectiveness in our framework.
 
\smallskip
\noindent \textbf{Our Contribution}.
%comment
\mycomment{
Many works in literature employ generative models such as GPT-2 and GPT-3 
%LLMs 
for rewriting conversational utterances by deriving term importance estimators or tracking the context's evolution through the conversation.}
%This work contributes to this line by exploiting and assessing the power of instructed LLM specifically trained for conversations, 
This work contributes to the line of rewriting conversational utterances with generative models. Differently from previous works, we assess the capabilities of an instructed LLM
such as ChatGPT in rewriting utterances after few-shot training. We experiment with different prompts and instructions to offer the model different amounts and kinds of information for obtaining utterance rewritings that are competitive with---or better than---the state-of-the-art.
%We perform the evaluation with reproducible experiments based on a publicly-available TREC CAsT datasets. %Then, we propose and evaluate different prompting approaches for deriving effective rewritings.

In this study, we evaluate ChatGPT's performance in explicit utterance rewriting. We conduct a comparative analysis with other state-of-the-art models employing explicit rewriting techniques \cite{10.1145/3397271.3401130,10.1145/3397271.3401323}. We acknowledge the potential contribution of dense retrieval approaches for utterance rewriting, as they can be applied after explicitly rewriting utterances. These approaches will be assessed in future research. % following the refinement of the explicit rewriting phase.

%\fnote{aggiungere che facciamo riscrittura esplicita, non dense. la parte dense, a valle di riscrittura esplicita, future work. Cris}
% !TEX root = ../paper.tex
% !TeX spellcheck = en_US
\vspace{-1mm}
\section{Methodology}
\label{sec:methodology}
    %\vspace{-2mm}

Our goal is to understand %whether instructed LLMs are suitable for  rewriting conversational utterances and 
to which extent  a state-of-the-art instructed LLM can be used to improve conversational search effectiveness. To this respect, this work  assesses with reproducible experiments the rewriting capabilities of ChatGPT (RQ1) and investigates the impact of different prompts and instructions on the effectiveness of a two-stage conversational search pipeline  
(RQ2).
% \guido{We define a multi-turn conversation , i.e., $u_1, \ldots, u_N$. Let $u_i \in \mathcal{U}$ be the current utterance, while $u_1, \ldots,$ $u_{i-1}$ denote the previous utterances of the same conversation. From now on, the utterances $u_1, \ldots,$ $u_{i-1}$ of the conversation $\mathcal{U}$ will be referred to as \textit{CONTEXT}.
In Table \ref{tab:glossary}, we introduce the notation used to describe our task. Our rewriting system $\Theta$, based on an instructed LLM, can take as input many of the elements described in the table in order to perform the rewriting of the current utterance $u_i$ into a rewritten version $\hat{u}_i$.
More formally, a typical rewriting request consists of the following:

\begin{equation}
\label{eq:request}
    \Theta(s, \mathcal{E}, \mathcal{C}, p, u_i) = \hat{u}_i,
\end{equation}

where $s$ represents the scope, i.e., the general task instructions of how we want the system to behave, $\mathcal{E}$ is a conversation example different from the current one, $\mathcal{C}$ is the context of  $u_i$, and $p$ is the  prompt accompanying $u_i$, which explicitly instructs $\Theta$ detailing the rewriting request by adding specific desired characteristics, e.g., ``concise'', ``verbose'', and ``self-explanatory''.

%%%%%%%% glossary %%%%%%%%
\begin{table}
\centering
\caption{Notation.}\label{tab:glossary}
\begin{tabular}{{p{0.2\linewidth}p{0.7\linewidth}}}
\toprule
Symbol & Definition \\
\midrule
$\mathcal{U}$                 & A multi-turn conversation composed of a sequence of utterances asked by a user to a conversational assistant. \\

$\Theta$               & An instructed LLM that we use for utterance rewriting, also referred to as  \emph{Assistant}. \\

$u_i$              & The current original utterance at turn $i$ in $\mathcal{U}$. \\

$\hat{u}_i$               & The current  utterance rewritten by $\Theta$. \\

$u_1, \ldots,$ $u_{i-1}$               & The previous original utterances in $\mathcal{U}$. \\

$\hat{u}_1, \ldots,$ $\hat{u}_{i-1}$ & The previous  utterances in $\mathcal{U}$ rewritten by $\Theta$. \\

$\bar{u}_1, \ldots,$ $\bar{u}_{i-1}$ & The previous manually-rewritten utterances in $\mathcal{U}$.\\

%$r_1, \ldots,$ $r_{i-1}$ & The canonical responses corresponding to the previous utterances. \\

$\hat{r}_1, \ldots,$ $\hat{r}_{i-1}$ &  Responses to the previous utterances generated by $\Theta$.\\

$\mathcal{C}$ & The \emph{Context} which is composed of the alternation between $u_1, \ldots,$ $u_{i-1}$ and $\hat{u}_1, \ldots,$ $\hat{u}_{i-1}$, or even adding
%$r_1, \ldots,$ $r_{i-1}$ or 
$\hat{r}_1, \ldots,$ $\hat{r}_{i-1}$. An example can be seen in Figure \ref{fig:diagramma}. \\

$\mathcal{E}$               & The \emph{Example} comprises original utterances $u_1, \ldots,$ $u_{i-1}$ and their corresponding manually rewritten utterances $\bar{u}_1, \ldots,$ $\bar{u}_{i-1}$. \\

$s$             & The scope that explains our goal to the rewriting LLM $\Theta$, also referred to as \emph{System}. \\

$p$  & The actual Prompt that, given $u_i$, specifies the instruction to $\Theta$, namely, to rewrite the query.\\

\bottomrule
\end{tabular}
\vspace{-5mm}
\end{table}
%%%%%%%% end glossary %%%%%%%%

%In Section \ref{subsec:prompts}, we  go into detail about how we construct the input for $\Theta$ and we describe the possible combinations of the elements used and actual prompts used to produce the rewriting.

\subsection{Instructed LLM}
\label{subsec:llm}

We employ ChatGPT as the instantiation of $\Theta$. Specifically, we employ the \gpt model. As indicated in the ChatGPT API description\footnote{\url{https://platform.openai.com/docs/guides/chat/instructing-chat-models}}, the model takes ``a series of messages as input and returns a model-generated message as output''. %However, there are several models available at the time which could be further investigated. According to OpenAI,
%\elnara{
%We use the free gpt-3.5-turbo API\footnote{https://platform.openai.com/docs/models/gpt-3-5} of the GPT-3.5 model family for contextual learning rewrites. 
Since the model does not provide memory or session retention, in each interaction, we enclose the interaction history of previous turns of the conversation into the current request. This leads to having a conversational-style request, similar to an actual dialog.

We adapt our utterance rewriting requests to 
the input structure of the \gpt model. The requests are composed of three main elements: \textit{system}, \textit{user}, and \textit{assistant}.  The ``system'' content is provided at the start of the session to specify the scope of the following interactions, in our case $s$. The ``user'' and ``assistant'', on the other hand, indicate the interactions between the user and ChatGPT, as a series of user instructions/requests consisting of prompt and current original utterance $(p, u_i)$, and the corresponding assistant response containing the rewritten utterance $\hat{u}_i$.

%\guido{From now on the three different inputs will be respectively referred to as \textit{system input,} \textit{assistant input} and \textit{user input} \fnote{really???}. Also, we will refer to the last user input, the one to which the model is supposed to answer, as \textit{instruction}.}
%\cris{AI assistant} \inote{chatgpt}?.

To better understand what the best way of prompting the system is, we experiment with different ways of providing ChatGPT with the prompt and  the context.

\begin{table*}[t]
    \caption{Example of conversation taken from the \castXIX evaluation dataset. The ``qid'' column identifies the conversational turn, the first number refers to the conversation while the second to the dialogical turn. The ``Raw utterance'' column reports the original utterance, the ``Manually-rewritten utterance'' column reports the human assessor rewritings, and the ``Automatically-rewritten utterance'' column reports an example of rewriting generated by ChatGPT with our proposed prompt P5.    \label{tab:conversationexample}}
    \centering
    \begin{tabular}{p{0.6cm}|p{4cm}|p{5.8cm}|p{5.8cm}}
    %{l|l|l|l}
    \toprule
    \textbf{qid}  & \textbf{Raw utterance} & \textbf{Manually-rewritten utterance} & \textbf{Automatically-rewritten utterance}\\\midrule
    31\_1 & What is throat cancer? & What is throat cancer? & What is throat cancer?\\
    31\_2 & Is it treatable? & Is throat cancer treatable? & Is throat cancer treatable?\\
    31\_3 & Tell me about lung cancer. & Tell me about lung cancer. & Tell me about throat cancer.\\
    31\_4 & What are its symptoms?  & What are lung cancer's symptoms? & What are the symptoms of throat cancer?\\
    31\_5 & Can it spread to the throat? & Can lung cancer spread to the throat? & Can throat cancer spread to other areas of the throat?\\
    31\_6 & What causes throat cancer? & What causes throat cancer? & What causes throat cancer to spread to other areas of the throat?\\
    31\_7 & What is the first sign of it? & What is the first sign of throat cancer? & What are the early signs of throat cancer?\\
    31\_8 & Is it the same as esophageal cancer? & Is throat cancer the same as esophageal cancer? & Is throat cancer the same as esophageal cancer? \\
    31\_9 & What's the difference in their symptoms? & What's the difference in throat cancer and esophageal cancer's symptoms? & What are the differences in the symptoms of esophageal cancer and throat cancer? \\
    \bottomrule
    \end{tabular}
    \vspace{-5mm}
\end{table*}

\subsection{Prompting ChatGPT}
\label{subsec:prompts}

%To answer our research questions,
We present \promptnumber different prompts $p$ to ask the instructed LLM to rewrite the utterances of a conversation $\mathcal{U}$. % by providing them with the original utterance $u_i$ and the \emph{Context} of the conversation up to that point. 

The typical request submitted through the ChatGPT APIs\footnote{\url{https://platform.openai.com/docs/api-reference}} contains the elements detailed in Eq. \ref{eq:request}, namely, scope, example, context, prompt, and current utterance. For all \promptnumber prompts the example $\mathcal{E}$ consists of an exemplary conversation, chosen randomly from the dataset and not related with $\mathcal{U}$, where  the user inputs are the original utterances, and the assistant inputs are instead the same utterances rewritten manually. Moreover, the context $\mathcal{C}$ consists of the previous utterances of $\mathcal{U}$, where the user inputs are the original utterances $u_1, \ldots,$ $u_{i-1}$, and the assistant inputs are instead the same utterances rewritten by the model, $\hat{u}_1, \ldots,$ $\hat{u}_{i-1}$. The only exception to this request template is \guido{the prompt} P1, where the context consists of the previous utterances $u_1, \ldots,$ $u_{i-1}$, and the assistant inputs consist of the previous utterances rewritten by the model $\hat{u}_1, \ldots,$ $\hat{u}_{i-1}$ together with the generated answers $\hat{r}_1, \ldots,$ $\hat{r}_{i-1}$.

\begin{figure}[ht]
    \centering
    \includegraphics[width=\linewidth]{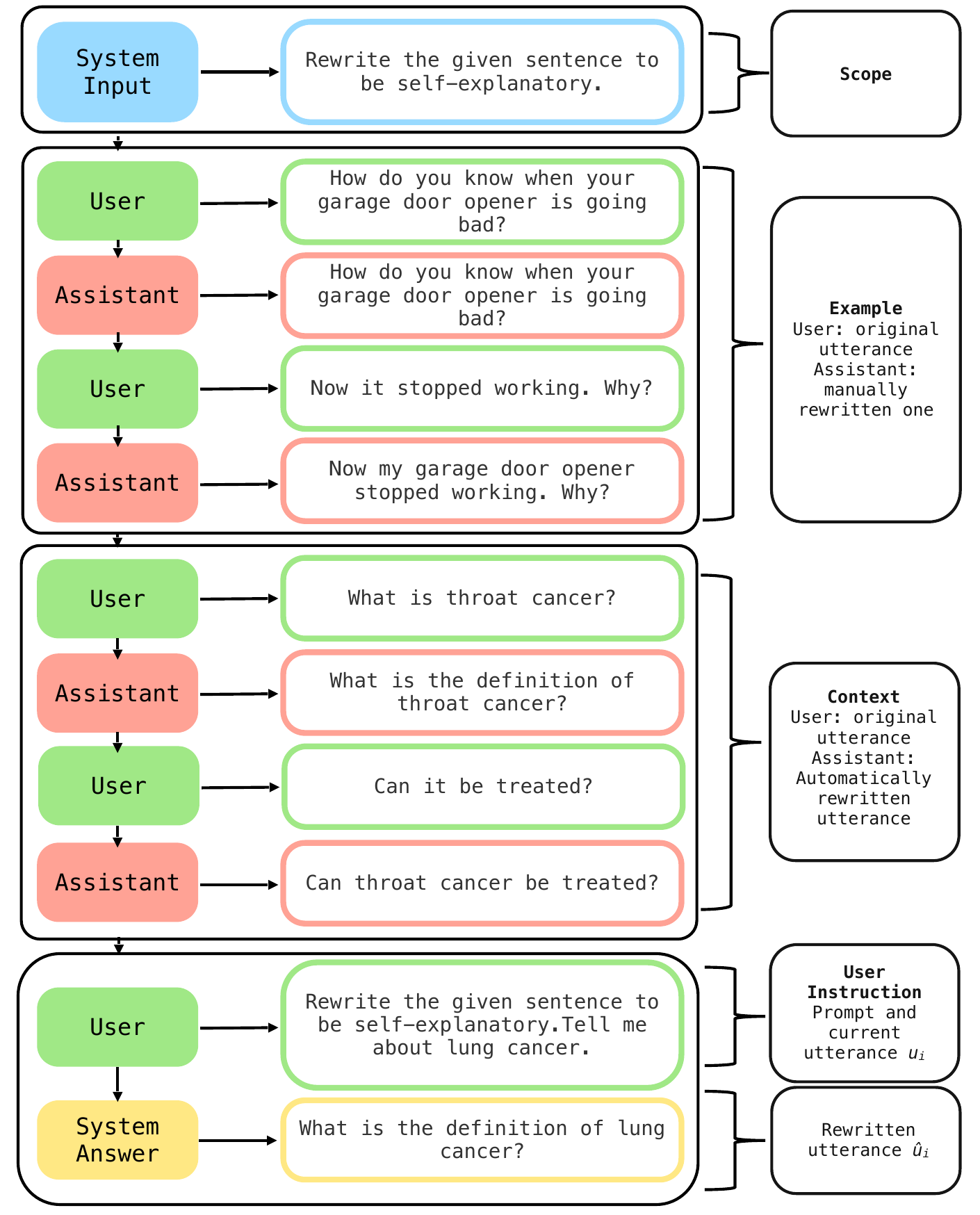}
    \caption{Main elements of an utterance rewriting request. \guido{The \textit{Scope} indicates the task that the model should perform. The \textit{Example} is the artificial part of the interaction where the user part is the query to rewrite and the assistant part is the query rewritten by a human. The \textit{Context} is composed of the previous queries rewritten by our model. The last section represents the current prompt and the output of the system.}}  
    \label{fig:diagramma}
    \vspace{-5mm}
\end{figure}

In Figure \ref{fig:diagramma}, we report a visual example of a typical rewriting request.  %Specifically, the prompt P7 with the best configuration C. 
We can see how the first block represents the system, the second one is the example and the third is the context of the current conversation, while the last one contains the prompt and current question, followed by the answer (rewritten utterance) provided by the assistant.

The above structure of the requests allows us to assess not only the rewriting capabilities of the model but also its  proficiency in retaining and exploiting the context information fed to the system.
We now detail the \promptnumber prompts with their specific characteristics and the intuition behind each of them.
%}the notation detailed in Equation \ref{ in Figure \ref{fig:diagramma}, the request template contains:

\iffalse
\begin{itemize}
    \item $s$ is the same as $p$ but it marks the beginning of the request.
    \item $\mathcal{E}$ \textit{Example:} the user inputs are the previous utterances, $u_1, \ldots,$ $u_{i-1}$, the assistant inputs are the manually rewritten utterances, $ \bar{u}_1, \ldots,$ $\bar{u}_{i-1}$.
    \item $\mathcal{C}$ \textit{Context:} the user inputs are the previous utterances, $u_1, \ldots,$ $u_{i-1}$, the assistant inputs are the previous utterances rewritten by the model, $\hat{u}_1, \ldots,$ $\hat{u}_{i-1}$.
    \item $p$
    \item $u_i$
\end{itemize}
\fi

\begin{itemize}[leftmargin=*,itemsep=0mm, parsep=0pt]

    \mycomment{\item[P1] \textit{Prompt:} ``Rewrite the following question to increase retrieval metrics of an information retrieval system. You can use previous questions to add valuable information. Return only the rewritten question.'' \\ 
    % \textit{Context:} $\mathcal{C}(u_1, \ldots,$ $u_{i-1}$, $\hat{u}_1, \ldots,$ $\hat{u}_{i-1})$.\\ 
    % \inote{I can make this a list of tuple $[(u_1, \hat{u}_1),\ldots, (u_{i-1},\hat{u}_{i-1})]$}\\
    \textit{Rationale:} P1 aims to specify the final goal of the rewriting while keeping track of the context to see if the model is able to maximize the retrieval results.}

    %used to be P2
    \item[P1] \textit{Prompt:} ``Rewrite the following question to be clear and complete and then provide an answer. Use the previous questions and answers to rewrite the question.''\\ 
%    \textit{Context:} the user inputs are the previous utterances $u_1, \ldots,$ $u_{i-1}$, and the assistant inputs are the previous utterances rewritten by the model together with the answer, $\hat{u}_1, \ldots,$ $\hat{u}_{i-1}$ and $\hat{r}_1, \ldots,$ $\hat{r}_{i-1}$.\\
    \textit{Rationale:} P1 aims to instruct the model to generate a self-explanatory sentence using not only the information provided by the previous utterances but also by the generated answers.

    %used to be P3
    \item[P2] \textit{Prompt:} ``Rewrite the following question adding keywords for a retrieval system. Use the information from the previous questions. Return only the rewritten question.'' \\
    \textit{Rationale:} P2 aims to specify the final goal of the rewriting while keeping track of the context to see if the model is able to maximize the retrieval results.
    %P2 is a slight variation of P1, but as we will see in Section \ref{sec:results}, the results obtained by the two differ significantly. The aim was to evaluate whether similar prompts could obtain different results.

    \mycomment{
    \item[P4] \textit{Prompt:} ``Rewrite the question to be clearer and more concise and provide as much context and information as possible to ensure that the proper context is maintained. Save the context from the previous question. Do not add any extra sentences.''\\
    \textit{Rationale:} P4 aims at retaining the contextual information given by the previous utterances to rewrite a self-explanatory sentence. A warning to avoid extra sentences is used because some responses added superfluous phrases in which the assistant asked to clarify the topic of the question.}

    %used to be P5
    \item[P3]  \textit{Prompt:} ``Rephrase the current question into a more concise and context-free form that is suitable for a multi-turn information search dialog using the context of the previous question. Do not add any extra sentences or notes.'' \\
    \textit{Rationale:} P3 aims to specify the final goal of the rewriting in the prompt and to instruct the model to generate a complete and concise rewriting of the given utterance.

    %used to be P6
    \item[P4] \textit{Prompt:} ``Reformulate the current question following the examples. [a list of 8 example pairs where each pair has the format ``\textit{Question: } raw question. \textit{Rewritten:} manually rewritten question''].'' \\
    \textit{Rationale:} P4 aims at reproducing the pattern given in the prompt to better rewrite the given utterances. Besides providing the example $\mathcal{E}$ within the request, we also repeat it in the prompt.
    
    %used to be P7
    \item[P5] \textit{Prompt:} ``In a multi-turn dialog system, rewrite the given sentence to be self-explanatory following the pattern of the previous interactions.'' \\ 
    \textit{Rationale:} P5 aims at reproducing the pattern given by the previous interactions between the user and the model, assuming that they are proficient in the rewriting task. %This prompt was specifically designed for the case in which part of the previous interactions is synthetic [a list of 5 exemplary interaction pairs where each pair has the format user: "\textit raw question.", assistant:  "\textit rewritten question"]]
%    \item[P7]Reformulate the current question into a de-contextualized rewrite under the multi-turn information-seeking dialog context. Then generate a correct response. Print also the reformulated question.
\end{itemize}

Moreover, we experiment in our setting also the best-performing prompt presented in the work of Mao \emph{et al.}~\cite{mao2023large}.

\begin{itemize}%[itemsep=1mm, parsep=0pt]  
    \item[E] ``Reformulate the current question into a de-contextualized rewrite under the multi-turn information-seeking dialog context. Then generate a correct response. Print also the reformulated question.''
\end{itemize}

We use the prompts above to generate rewritten utterances and test their effectiveness. We rewrite all the utterances of a conversation except the first one, $u_1$. In fact, several studies have shown that the first utterance of each conversation is already a self-explanatory sentence \cite{mele_adaptive_2021}.
%Keeping the original first utterance unchanged and using the \promptnumber prompts for rewriting the rest of the conversation %.providing the synthetic conversation between ChatGPT and the user shows 
%provides the best results.
%, for this reason from this point every result refers to this specific configuration:

Before selecting the \promptnumber prompts, we tested several other configurations not reported for the sake of brevity but resulting in worse performance. For example, we tried to use the prompt $p$ only as system input and not in every user input $u_i$. We also tested prompts not providing rewriting examples or using different textual instructions. %The results of these particular settings are not reported here for the sake of space because they showed worse results than the ones we discuss in the following sections.  
%We also tried to insert the context, $u_1, \ldots,$ $u_{i-1}$, at the end of each instruction. 
As a general consideration, we notice that explaining the input to ChatGPT in a detailed way (e.g., by specifying \textit{``In some cases, I will provide the questions previously made by the user. Use them to better reformulate the question.''}) improves the performance and avoids some rewritings errors. 
Finally, we observe that the output of ChatGPT sometimes contains additional elements (e.g., clarifying questions) or it directly includes  an answer to the utterance. For this reason, we post-process the output and keep only the actual rewritten utterance, $\hat{u}_i$.

%To obtain good quality utterance rewritings, we provide  $\Theta$ with the proper instructions and all the necessary information to make the utterances more clear and context-free compared to the original ones. 

\section{Experimental Setup}
\label{sec:expsetup}

To assess the utterance rewriting quality, we submit $\hat{u}_i$ as a query to a two-stage information retrieval pipeline. We evaluate the effectiveness of the different rewriting strategies using the TREC CAsT framework \cite{dalton_cast_2019,dalton_cast_2020,dalton_trec_2021}, which allows us to perform an objective evaluation by comparing our results to those obtained by state-of-the-art competitors\footnote{We will release the code used for the experiments and the full set of rewritten utterances tested to favor the  reproducibility of results.}.

\subsection{Conversational Datasets}
\label{subsec:datasets}
Our experiments are based on the TREC Conversational Assistant Track (CAsT) 2019 and 2020\footnote{Conversational Assistant Track, \url{https://www.treccast.ai/}} datasets. 
The \castXIX~\cite{dalton_cast_2019} dataset consists of 20 human-assessed test conversations, while \castXX~\cite{dalton_cast_2020} includes 25 conversations, with an average of 10 turns per conversation. The \castXIX and 2020 datasets include relevance judgments at the passage level. Conversations are provided with original and manually-rewritten utterances. %Original utterances are the raw utterances. 
The manually-rewritten utterances are the same conversational utterances as the original ones, where human assessors resolve missing keywords or references to previous topics. Relevance judgments have a three-point graded scale and refer to passages of the TREC CAR \guido{(TREC Complex Answer Retrieval)}, the MS-MARCO (MAchine Reading COmprehension) and the  WaPo (TREC Washington Post Corpus) collections for \castXIX and 2020 \guido{for a total of 38,636,520 passage}. %\fnote{Manca un pezzo secondo me}
%, and to documents of MS-MARCO, KILT Wikipedia, and Washington Post 2020 for \castXXI.
In these datasets, questions within a conversation are characterized by anaphora and ellipses. They imply a big part of the context and miss explicit references to the current topic. Table \ref{tab:conversationexample} reports some examples of utterances from the \castXIX dataset. We can see that manually-rewritten utterances are concise and rephrase the original utterance by adding the missing tokens to make it self-explanatory. On the other hand, depending on the prompt, automatically-rewritten utterances tend to be more verbose although well-formed natural language questions.

%Our experiments are based on 2019, and 2021, and 2021 TREC Conversational Assistant Track (CAsT)\footnote{Conversational Assistant Track, \url{https://www.treccast.ai/}} datasets. The \castXIX~\cite{dalton_cast_2019} dataset consists of 20 human-assessed test conversations, while \castXX~\cite{dalton_cast_2020} and \castXXI includes 25 and 26 conversations respectively, with an average of 10 turns per conversation. The \castXIX and 2020 include relevance judgments at passage level, whereas for CAsT 2021 the relevance judgments are provided at the document level. The judgments have a three-point graded scale and refer to passages of the TREC \ac{CAR}, and MS-MARCO (MAchine Reading COmprehension) collections for \castXIX and 2020, and to documents of MS-MARCO, KILT Wikipedia, and Washington Post 2020 for \castXXI.
%In our experiments we try to predict nDCG@3, the most commonly used measure in \ac{CS}~\cite{DaltonEtAl2020,DaltonEtAl2020b}.
% CAST 2019: 50 test topics, from 7 to 12 utt each, 9.58 utt/conv on avg
% CAST 2020: 25 topics, from 6 t 13, 8.64
% CAST 2021: 26 topics, from 6 to 13, 9.192

%%% INSTRUCTED E PROMPTS ERA QUI
%\fnote{Vecchia descrizione della figura}
    \vspace{-1mm}

\subsection{Baselines}
\label{subsec:baselines}
We assess the retrieval effectiveness 
%in terms of Reciprocal Rank, P@1, NDCG@3, R@200 
of original, manually-rewritten, and automatically-rewritten utterances.
% Original utterances are the raw utterances provided in the TREC CAsT datasets. In contrast, manually-rewritten queries are the same conversational queries, where human assessors resolve missing keywords or references to previous topics, while with automatically-rewritten queries, we indicate the conversational queries rewritten by an automatic technique.
%that includes: i) the ones obtained with our instructed LLM prompting methods along with the one by Mao , ii) two competitor methods such as CQR %cite{10.1145/3397271.3401323} 
%or QuReTec. %~\cite{10.1145/3397271.3401130}.
In detail, we consider %experimentally evaluate our proposed rewriting approaches by comparing against 
the following rewriting methods and baselines:
\begin{itemize}%[itemsep=1mm, parsep=0pt]
    \item \textit{Original utterances}: raw utterances provided by TREC CAsT. % produced by a user, in a non formal way, to query the system.
    %often characterized by anaphora or ellipsis, in which the context is implicit rather than explicit.
    \item \textit{Manual utterances}: manually-rewritten utterances by human annotators
    %to explicitly mention the context from the conversation and make the questions self-explanatory. 
    provided by TREC CAsT. 
    \item \textit{QuReTeC}~\cite{10.1145/3397271.3401130}: utterances %that
    are rewritten with a BiLSTM sequence to sequence model trained for query resolution.
    \item \textit{CQR self-learn cv}~\cite{10.1145/3397271.3401323}: utterances are generated in two steps, first with a GPT-2 model trained with self-supervised learning to generate contextual utterances containing few information presented in previous utterances. The second step is performed with a GPT-2 model fine-tuned on manual rewrites via five-fold cross-validation.
    \item \textit{CQR rule-based cv}~\cite{10.1145/3397271.3401323}: utterances are generated in two steps, first with a rule-base approach that deals with omissions and coreference and successively rewritten with a GPT-2 model fine-tuned on manual rewrites via five-fold cross-validation.
    \item \textit{Prompt E}~\cite{mao2023large}: although the results by Mao \emph{et al.} are achieved on a different generative model, i.e., GPT-3, we use their prompt in our experimental framework to compare its retrieval performance against ours.
\end{itemize}

\subsection{Two-stage Retrieval}
\label{subsec:twostage}
To evaluate and compare the different utterance rewritings, we index the TREC CAsT collections by removing stopwords and applying Porter’s English stemmer. 
%We use the original, manually, and automatically-rewritten utterances as queries. 
We use PyTerrier~\cite{macdonald_pyterrier_2021} to build the information retrieval pipeline, which is composed of two stages:
\begin{itemize}%[itemsep=1mm, parsep=0pt]
    \item The first stage performs document retrieval on the indexed collection with the DPH weighting model~\cite{10.1145/582415.582416}, using the raw, manually, and automatically-rewritten utterances; %as queries;
    \item The second stage performs reranking of the top-$1000$ candidates retrieved by the first stage by using the MonoT5 model~\cite{Pradeep2021TheED} made available in PyTerrier\footnote{\url{https://github.com/terrierteam/pyterrier\_t5}}.
\end{itemize} 

%\fnote{non diciamo se facciamo stemming, ecc..}

We measure the retrieval effectiveness of the first stage and of the second stage using the following metrics: Mean Reciprocal Rank (MRR), Precision@1 (P@1), Normalized Discounted Cumulative Gain@3 (NDCG@3), and Recall@500 (R@500).
MRR and NDCG@3 are standard metrics used for evaluation purposes in the TREC CAsT framework while the others are included to provide a more comprehensive evaluation of the retrieval capabilities of the first-stage (R@500) and the reranking capabilities of the second-stage (P@1).
%\fnote{aggiungerei frasetta dicendo che sono le metriche usate nel framework cast. FM}

%%%%%%%%%%%% DENSE RETRIEVAL %%%%%%%%%%%%%%%%

% Furthermore, we also evaluated the top-performing generated rewritings with ANCE \cite{ance} embeddings  CONVDR\cite{yu_few-shot_2020} embeddings

% \fnote{ La parte qui sotto non so quanto sia giusta, dovrebbe dare un'occhiata Cris}\fnote{infatti e sbagliato}

% \cris{Original and manually rewritten queries, as well as all the automatically rewritten ones are encoded using the ANCE\footnote{https://huggingface.co/sentence-transformers/msmarco-roberta-base-ance-firstp} model\cite{ance-paper}. For dense retrieval we also include the ConvDR\cite{10.1145/3397271.3401323} baseline, which is based on ANCE but fine-tunes embeddings using a teacher-student model, so as to make the embeddings more similar to the manual ones. For ConvDR we used publicly available weights\footnote{https://github.com/thunlp/ConvDR}.}

% \cris{In the dense retrieval experiments, based on dense representations, we encode both documents and queries using the ANCE model, resulting in 768-dimensional embeddings. All the document collection embeddings are indexed in a Flat Index, based on the FAISS library. that allows exact results. For retrieval the search performs KNN using Inner Product between the query and documents vectors.}

 %Notice that we do not report ConvDR results for the \castXXI since, at the current time, there are no publicly available weights for this dataset.

% !TEX root = ../paper.tex
% !TeX spellcheck = en_US

\begin{table*}[t]
    \caption{First-stage retrieval results in terms of MRR, P@1, NDCG@3 and R@500 on \castXIX and \castXX datasets. In bold, we report the best results achieved for each metric, except Manual. We mark statistically-significant performance gain/loss, calculated with the two-paired $t$-test ($p$-value $< 0.05$) with Bonferroni correction, of our methods with respect to the QuReTeC and CQR self-learn cv baselines with the symbols $\blacktriangle$ and $\blacktriangledown$ for the first, $\triangle$ and $\triangledown$ for the latter.\label{table:first-stage}}
    \centering
    \renewcommand{\arraystretch}{1.1}
    \begin{tabular}{c|cccc|cccc}
    \toprule
        ~ & \multicolumn{4}{c}{\textbf{\castXIX}} & \multicolumn{4}{c}{\textbf{\castXX}} \\\midrule
        \textbf{Prompt} & \textbf{MRR} & \textbf{P@1} & \textbf{NDCG@3} & \textbf{R@500} & \textbf{MRR} & \textbf{P@1} & \textbf{NDCG@3} & \textbf{R@500} \\
        \midrule
        Manual & 0.6753$\triangle$ & 0.5491 & 0.4002$\triangle$ & 0.7374$\triangle$$\blacktriangle$ & 0.6220$\blacktriangle$ & 0.5048$\blacktriangle$ & 0.3277$\blacktriangle$ & 0.6682$\blacktriangle$ \\
        Original & 0.3334$\triangledown$$\blacktriangledown$ & 0.2254$\triangledown$$\blacktriangledown$ & 0.1617$\triangledown$$\blacktriangledown$ & 0.3815$\triangledown$$\blacktriangledown$ &  0.2177$\blacktriangledown$ & 0.1587$\blacktriangledown$ & 0.0998$\blacktriangledown$ & 0.2532$\blacktriangledown$ \\
        \midrule         
%        P1 & 0.5587 & 0.4104 & 0.2809 & 0.4317 $\triangledown$$\blacktriangledown$ & 0.4359	& 0.3462 & 0.2038 & 0.3800\\ 
        P1 & 0.6327 & \textbf{0.5260} & \textbf{0.3664} & 0.6446 & \textbf{0.5353}$\blacktriangle$ & \textbf{0.4231}$\blacktriangle$ & \textbf{0.2512} & \textbf{0.5710}\\ 
        P2 & 0.5887 & 0.4624 & 0.2921 & 0.5775$\triangledown$$\blacktriangledown$  & 0.4838 & 0.3750 & 0.2406 & 0.5488\\ 
%        P4 & 0.5478 & 0.4046 & 0.2848 & 0.4349$\triangledown$$\blacktriangledown$ & 0.4539 & 0.3510 & 0.2073 & 0.3950\\ 
        P3 & 0.6129 & 0.5087 & 0.3363 & 0.6036$\blacktriangledown$ & 0.4580 & 0.3150 & 0.2153 & 0.5009\\ 
        P4 & 0.6221 & 0.5116 & 0.3449 & 0.6311 & 0.4302 & 0.3317 & 0.2109 & 0.4963\\ 
        P5 & \textbf{0.6359	}& 0.5145 & 0.3331	& 0.6499	& 0.4775 & 0.3894 & 0.2266 &	0.5133 \\ 
        \midrule
        E & 0.5837 & 0.4798 & 0.3094 & 0.5772$\blacktriangledown$  & 0.4520 & 0.3558 & 0.2181 & 0.5029\\ \midrule
        QuReTec\cite{10.1145/3397271.3401130} & 0.6251 & 0.4913 & 0.3494 & \textbf{0.6704}  & 0.4399 & 0.3221 & 0.2145 & 0.5163 \\ 
        CQR self-learn cv\cite{10.1145/3397271.3401323} & 0.5915 & 0.4682 & 0.3336 & 0.6617 & - & - & - & -  \\ 
        CQR rule-based cv\cite{10.1145/3397271.3401323} & 0.5629 & 0.4162 & 0.3111 & 0.6569 & - & - & - & -\\
        \bottomrule
    \end{tabular}
    \vspace{-5mm}
\end{table*}

\section{Results and Discussion}
\label{sec:results}
In this section, we discuss the experimental results on \castXIX and 2020 datasets  to assess the various rewriting strategies and compare them with the baselines.
%The code and the data used to run the experiments is available for permitting the reproduction of our results\footnote{The link will be added upon acceptance of the paper.}.

\subsection{First-stage Retrieval}
In Table \ref{table:first-stage}, we report the results obtained when performing document retrieval using the DPH weighting model\cite{10.1145/582415.582416}. Results refer to the first-stage retrieval pipeline on both the \castXIX and \castXX datasets. We also experiment with other weighting models, i.e., BM25\cite{10.1561/1500000019}. We do not report them as their results are worse than those achieved by DPH.\\
The performance of our methods and baselines range between the ones obtained for the original and the manually-rewritten utterances. 
%The best-performing method for \castXIX in terms of P@1, which is one of the most indicative metrics for conversational search, is the one based on P7 with $0$.$5145$, while in terms of NDCG@3, QuReTec takes the lead with $0$.$3494$.
Considering \castXIX,  P5 is the best-performing prompt when looking at MRR while P1 is the best-performing prompt in terms of Precision@1 and NDCG@3. Regarding R@500, the QuReTec baseline is the best-performing method.
When performing the statistical significance evaluation using a two-paired $t$-test ($p$-value $< 0.05$) with the Bonferroni correction~\cite{bonferroni}, the results achieved by our prompts are not  statistically different from the state-of-the-art baselines, except for R@500 for P2, P3, and E.\\
%It is worth noticing that this result is mostly due to the fact that we have a small number of queries with relevance labels.
Improved results are achieved when rewriting the utterances of the \castXX evaluation dataset. The best-performing rewriting method is based on P1, where all metrics show considerable gains over the QuReTec baseline. For P@1 and MRR,  the improvement achieved by P1 is statistically significant when compared to the QuReTec baseline, with a $21$.$6$\% gain in MRR and $31$.$7$\% in P@1. NDCG@3 and R@500 increase by $17$.$1$\% and $10$.$6$\%, respectively. We remind the reader that P1 also considers the generated answers to the previously rewritten questions to produce the current rewriting. In fact, it is worth noting that, compared to \castXIX where most relevant concepts could be found in the previous utterances, for \castXX, some missing relevant concepts that fill out the context, can be found only in the responses and not in the utterance history. Results show that by generating the answers to the user requests and instructing the model to use them in the rewriting phase, we obtain improved results.\\
The fact that, independently of the dataset considered, our few-shot rewriting system obtains results as good as---or better than---state-of-the-art techniques should be further exploited in future work.

%In this case, even though we managed to get interesting results we are not able to achieve results as good as the ones obtained when evaluating the manually rewritten queries or the selected baselines, namely CQR and QuReTec. 

%One consideration that must be done, is that the first-stage retrieval is performed using a bag-of-words approach, which means that the queries rewritten by the \gpt model might contain elements that do not help enough while retrieving with such an approach. In fact, analyzing the automatically-rewritten utterances we observed that in many cases the system produces verbose rewritings that seem both grammatical and clear, but that may add noise for a bag-of-word approach.

%Nonetheless, we obtained overall results, showing improvements over the competitor baseline baselines.

%\guido{Also, we notice that we obtain better results when rewriting the utterances of the \castXX evaluation dataset. In fact, we manage to achieve better results when comparing with the QuReTeC baseline, with a relative improvement for Reciprocal Ranking up to 21.6 \%, for Precision@1 up to 31.7 \% and 17.1\% and 13.0 \% respectively for NDCG@3 and R@500.
%Also, it is worth noting that the results for the P1 prompt for Reciprocal Ranking and Precision@1 are statistically significant according to the paired p-test calculated with the Bonferroni correction\cite{bonferroni}.
% }
%Nonetheless, we obtained overall good results and in some cases were able to be better than the ones of the baselines. 

\begin{table*}[t]
    \caption{Second-stage retrieval results in terms of MRR, P@1, NDCG@3 and R@500  on \castXIX and \castXX datasets. In bold, we report the best results achieved for each metric, except Manual. We mark statistically-significant performance gain/loss, calculated with the paired $t$-test ($p$-value $< 0.05$) with Bonferroni correction, 
    of our corresponding methods with respect to the QuReTeC and CQR self-learn cv baselines with the symbols $\blacktriangle$ and $\blacktriangledown$ for the first,  $\triangle$ and $\triangledown$ for the latter.\label{table:second-stage}} 
    \centering
    \renewcommand{\arraystretch}{1.1}
    \begin{tabular}{c|cccc|cccc}
    \toprule
        ~ & \multicolumn{4}{c}{\textbf{\castXIX}} &  \multicolumn{4}{c}{\textbf{\castXX}} \\\midrule
            \textbf{Prompt} & \textbf{MRR} & \textbf{P@1} & \textbf{NDCG@3} &\textbf{R@500} & \textbf{MRR} & \textbf{P@1} & \textbf{NDCG@3}&\textbf{R@500} \\
        \midrule
        Manual & 0.8849$\triangle$$\blacktriangle$  & 0.8266$\triangle$$\blacktriangle$ & 0.6053$\triangle$$\blacktriangle$ & 0.7705$\triangle$$\blacktriangle$ & 0.8161$\blacktriangle$ & 0.7308$\blacktriangle$ & 0.5381$\blacktriangle$ & 0.7361$\blacktriangle$  \\ 
        Original & 0.4643$\triangledown$$\blacktriangledown$ & 0.3989$\triangledown$$\blacktriangledown$ & 0.2791$\triangledown$$\blacktriangledown$ & 0.4060$\triangledown$$\blacktriangledown$ & 0.3301$\blacktriangledown$ & 0.2212$\blacktriangledown$ & 0.1813$\blacktriangledown$ & 0.2834$\blacktriangledown$\\
        \midrule
 %       P1 &  0.7840 & 0.6879 & 0.4826 & 0.6229$\blacktriangledown$ &  0.6288 & 0.5481 & 0.3785 & 0.5142\\ 
        P1 &  0.7909 & 0.6936 & 0.5193 & 0.6974  & \textbf{0.7249}$\blacktriangle$ &  \textbf{0.6394}$\blacktriangle$ & \textbf{0.4386}$\blacktriangle$ & \textbf{0.6287}$\blacktriangle$\\ 
        P2 & 0.7440 & 0.6358 & 0.4829$\blacktriangledown$ & 0.6347$\blacktriangledown$ & 0.6758 & 0.5962 & 0.4220$\blacktriangle$ & 0.6091\\ 
   %     P4 & 0.7415 & 0.6416 & 0.4862$\blacktriangledown$ & 0.6268$\blacktriangledown$ & 0.6581 & 0.5625 & 0.3902 & 0.5333\\ 
        P3 & 0.7377 & 0.6647 & 0.4867 & 0.6419$\blacktriangledown$ & 0.6022 & 0.5144 & 0.3542 & 0.5597\\ 
        P4 & 0.7575 & 0.6532 & 0.5155 & 0.6710 & 0.6086 & 0.5240 & 0.3601 & 0.5469  \\ 
        P5 & \textbf{0.8119} & \textbf{0.7283} & \textbf{0.5343} & 0.7059 & 0.6536 & 0.5721 & 0.4046 & 0.5650\\ \midrule
        E &  0.6863 & 0.5954 & 0.4507 & 0.6157$\blacktriangledown$ & 0.6163 & 0.5481 & 0.3855 & 0.5572\\ \midrule
        QuReTec\cite{10.1145/3397271.3401130} & 0.7858 & 0.6879 & 0.5330 & \textbf{0.7111} & 0.5788 & 0.4856 & 0.3454 & 0.5639 \\ 
        CQR self-learn cv\cite{10.1145/3397271.3401323} & 0.7780 & 0.7052 & 0.5286 & 0.6938  & - & - & - & -   \\ 
        CQR rule-based cv\cite{10.1145/3397271.3401323} & 0.7630 & 0.6821 & 0.5109 & 0.6853  & - & - & - & -  \\
        \bottomrule
    \end{tabular}
    \vspace{-5mm}
\end{table*}
\vspace{-1mm}
\subsection{Second-stage Retrieval}
%r what concerns the second stage of retrieval, we used the MonoT5 model implemented in PyTerrier to perform reranking on the queries generated by the \gpt model. 

In Table \ref{table:second-stage}, we report the end-to-end results obtained with \castXIX and 2020 %datasets
when performing document re-ranking using the MonoT5 model in the second-stage retrieval pipeline. 

Our intuition is that because our rewriting techniques produce verbose and well-formed utterance rewritings, it would be beneficial to use a LLM-based model such as T5, so as to effectively exploit the information added by the \gpt model. We can see that the performance obtained by the generated rewritings achieves higher results than those obtained by the CQR and QuReTec competitors for prompts such as P1, P5 for \castXIX, and for all prompts for \castXX.

The winning method for \castXIX is P5, with an MRR of $0$.$8119$ (3.3\%  increase), P@1 of $0$.$7283$ (5.9\%  increase), NDCG@3 of $0$.$5343$ that is slightly better than the one provided by QuRETec, i.e., $0$.$5330$. Consistent with the first stage, also in the second-stage retrieval, the results are better with respect to the QuReTec baseline, except for R@500, although not statistically significant.
%except P3 and P4 that show lower NDCG@3, and P1 and E with lower R@500.

When considering the \castXX evaluation dataset, our rewriting methods show significant improvements after reranking. In this case, we have a clear winner, i.e., P1, for which all metrics improve over QuReTec in a statistically-significant way. The MRR increases by $25$.$2$\%, the P@1 by $31$.$7$\%, the NDCG@3 by $27$.$0$\%, and the R@500 by $11$.$5$\%. Also, for P2, we have a statistically-significant improvement of $22$.$17$\% in terms of NDCG@3.

Even in the second stage of retrieval, we obtain results as good as---or better than---state-of-the-art competitors, confirming  that instructed  LLMs  are effective in rewriting utterances in a multi-turn conversational setting. %Furthermore, as observed before, the complexity of \castXX is higher than \castXIX, nonetheless, the model  generates rewritings of a good quality for retrieval purposes.

%It is important to note that the relative increase discussed above showed a statistical significance when compared with the QuReTeC baseline with the Bonferroni correction.
\vspace{-1mm}
\subsection{Answering our Research Questions} 
\noindent \textbf{RQ1}. We affirm that using an instructed LLM to rewrite utterances helps the effectiveness of the retrieval system. In fact, we can observe that for the \castXX dataset, we obtain significant improvements over the QuReTeC baseline, while for the \castXIX we achieve the same results, and in some cases, we outperform QuReTeC and the two CQR competitors.

%Furthermore, it is worth noticing that the two baselines, i.e., QuReTeC and CQR self-rule cv, were originally built on the \castXIX dataset, and for the data we have generated with QuReTeC, the results were not as good as for \castXIX.
%\fnote{sopra, non ho capito! FM}

The results achieved also show that, although the LLM has not been fine-tuned explicitly for utterance rewriting, it provides competitive results compared to the state of the art. This confirms the ability of these models to perform a variety of tasks via few-shot learning, thus lowering the effort needed for targeting novel tasks. In fact, custom-made models for utterance rewriting in conversational search, i.e., QuReTec, reach worse results on \castXX than an instructed LLM with well-designed prompts. We explain these results as a consequence of the capability of an LLM to deal with different datasets and domains, keeping a rewriting quality higher than other systems trained on limited data and thus characterized by a lower generalization power.

%\fnote{more or less? sopra, non ho capito. chatgpt e' allenata su task conversazionale. ha miliardi di parametri. che si vuole dire?}

\noindent \textbf{RQ2}.
For what concerns the best way of prompting the LLM, the best results are obtained with P1 for \castXX, while with P1 and P5 for \castXIX. 
While for some of the prompts discussed we clearly explicit the scope of the rewriting (e.g. ''[...]for a retrieval system[...]'' in P2), in both P1 and P5 this information is not explicit, suggesting that this kind of instruction is not useful to obtain better rewritings.
%We can see that both prompts are basic but clear and do not refer to the scope of the rewriting.

Moreover, in both cases, there is a clear indication of how to exploit examples and context from the previous interactions. The difference %between the two%
is that P1 explicitly asks the model to also add previously generated answers to the context and use all the information for generating the rewriting $\hat{u}_i$. This proved particularly effective in the case of \castXX. This could also be the reason why \guido{QuReTec} %the QuReTec %baseline
underperforms as, by design, it only focuses on the previous utterance and does not integrate the content of the answers for generating the rewriting. 
Therefore, after establishing the best-performing prompts and observing that they both make use of the context, we can conclude that providing examples can have a significant impact on the model's capabilities in performing the chosen task.
%As stated above, we tested \gpt in other settings that we do not report here for the sake of space, evidencing a clear gain in the quality of the results when providing task examples.}

%Finally, we can affirm that \gpt is effective in rewriting utterances. In fact, we achieved in most of the cases better results than the baselines. 
%Furthermore, it will be interesting to observe how the newly released models such as GPT-4~\cite{openai2023gpt4}, and probably soon, BARD~\cite{lewis2019bart} will deal with the same task. }
%\vspace{-2mm}

% !TEX root = ../paper.tex
% !TeX spellcheck = en_US
\vspace{-1mm}
\section{Conclusions and Future Work}
\label{sec:conclusion}

In this paper, we proposed several methods for using an instructed LLM for the conversational utterance rewriting task.

We focused on assessing if such type of model is suitable for this task and if it is competitive with the current state-of-the-art rewriting techniques, which use models specifically fine-tuned for the task. We also studied different prompting techniques to assess the most effective ways to instruct the model using $5$ prompt formulations.

We evaluate our proposals on the publicly-available TREC \castXIX and \castXX datasets. We provide a comprehensive experimental evaluation of our proposed \promptnumber ways of prompting the instructed LLM and state-of-the-art conversational rewriting baselines by assessing their retrieval effectiveness in a two-stage retrieval pipeline.

Experiments show that, in most cases, our proposed rewriting methods outperform the baselines. The largest gain is achieved for \castXX with increases in MRR by 25.2\%, in P@1 by 31.7\%, in NDCG@3 by 27.0\%, and in R@500 by 11.5\%. These results are obtained using prompt P1, in which the system is also required to consider previous answers when rewriting the current utterance. We can conclude that using an instructed LLM is beneficial for the utterance rewriting task in conversational search. These models %Furthermore, we want to stress that the goal of this work is to inspect the capabilities of newly hyped models to understand whether they 
can become a useful tool to further expand rewriting approaches and set new state-of-the-art standards.

\smallskip
\noindent \textbf{Future Work}. As future work, we are interested in studying how instructed LLMs can be used to generate synthetic data that can be exploited in other tasks of conversational search or even for enriching conversational datasets with weak supervision labels. The limited number of assessed conversations is in fact one of the main limitations in the conversational search domain. Moreover, we are interested in assessing the sensibility of prompting, i.e., how the utterance rewriting changes with respect to variations in the prompt and how it influences the retrieval performance, in a systematic and comprehensive way. %\guido{ The next step will be to test more LLM models also with the latest versions of TREC CAsT datasets, and experiment with the proposed approaches  in other conversational settings.}% where the performance gap between automatically generated queries and manually rewritten ones.}

\smallskip
\noindent \textbf{Acknowledgements}. Funding for this research has been provided by: PNRR - M4C2 - Investimento 1.3, Partenariato Esteso PE00000013 - ``FAIR - Future Artificial Intelligence Research'' - Spoke 1 ''Human-centered AI'' funded by the European Union (EU) under the NextGeneration EU programme;  the EU’s Horizon Europe research and innovation programme EFRA (Grant Agreement Number 101093026).  Views and opinions expressed are however those of the author(s) only and do not necessarily reflect those of the EU or European Commission-EU. Neither the EU nor the granting authority can be held responsible for them.
%by the Horizon Europe RIA ``Extreme Food Risk Analytics'' (EFRA), grant agreement n. 101093026

%, and by ``SoBigData.it – Strengthening the Italian RI for Social Mining and Big Data Analytics'' – funded by the European Union – NextGenerationEU – National Recovery and Resilience Plan (PNRR) Prot. IR0000013 – Avviso n. 3264 del 28/12/2021.
\vspace{-2mm}

\bibliographystyle{IEEEtran}
\bibliography{sample-base}

\end{document}